\documentclass{article}

\usepackage{arxiv}

\usepackage[utf8]{inputenc}   % allow utf-8 input
\usepackage[T1]{fontenc}      % use 8-bit T1 fonts
\usepackage{ifpdf}            % PDFLaTeX conditional (arXiv recommended)
\usepackage{hyperref}         % hyperlinks (auto-detects PDFLaTeX driver)
\usepackage{url}              % simple URL typesetting
\usepackage{booktabs}         % professional-quality tables
\usepackage{amsfonts}         % blackboard math symbols
\usepackage{amsmath}
\usepackage{amssymb}
\usepackage{nicefrac}         % compact symbols for 1/2, etc.
\usepackage{microtype}        % microtypography
\usepackage{graphicx}         % figure inclusion (auto-detects driver)
\usepackage{multirow}
\usepackage{enumitem}
\usepackage{natbib}
\usepackage{doi}

\renewcommand{\shorttitle}{Benchmarking ML and BiLSTM+Attention for Indonesian Sentiment}
\renewcommand{\undertitle}{}
\renewcommand{\headeright}{}
\hypersetup{
  pdftitle={Benchmarking Logistic Regression, SVM, and LightGBM Against BiLSTM
            with Attention for Sentiment Analysis on Indonesian Product Reviews},
  pdfsubject={cs.CL, cs.LG},
  pdfkeywords={Sentiment Analysis, PyCaret, BiLSTM, Attention Mechanism,
               Indonesian Product Reviews, Machine Learning},
  colorlinks=true,
  linkcolor=black,
  citecolor=black,
  urlcolor=black
}

\title{Benchmarking Logistic Regression, SVM, and LightGBM\\
Against BiLSTM with Attention for Sentiment Analysis\\
on Indonesian Product Reviews}

\author{%
  Razin Hafid Hamdi \\
  \texttt{razin.123450096@student.itera.ac.id} \\
  \small Institut Teknologi Sumatera \\
  \small Lampung Selatan, Indonesia
  \And
  Ivana Margareth Hutabarat \\
  \texttt{ivana.123410028@student.itera.ac.id} \\
  \small Institut Teknologi Sumatera \\
  \small Lampung Selatan, Indonesia
  \AND
  Hanna Gresia Sinaga \\
  \texttt{hanna.123450038@student.itera.ac.id} \\
  \small Institut Teknologi Sumatera \\
  \small Lampung Selatan, Indonesia
  \And
  Luluk Muthoharoh \\
  \texttt{luluk.muthoharoh@sd.itera.ac.id} \\
  \small Institut Teknologi Sumatera \\
  \small Lampung Selatan, Indonesia
  \AND
  Ardika Satria \\
  \texttt{ardika.satria@sd.itera.ac.id} \\
  \small Institut Teknologi Sumatera \\
  \small Lampung Selatan, Indonesia
  \And
  Martin C.T. Manullang \\
  \texttt{martin.manullang@if.itera.ac.id} \\
  \small Institut Teknologi Sumatera \\
  \small Lampung Selatan, Indonesia
}

\date{}

\begin{document}
\maketitle

% ---------------------------------------------------------------
% Abstract
% ---------------------------------------------------------------
\begin{abstract}
Sentiment analysis of product reviews on e-commerce platforms plays a critical role in automatically
understanding customer satisfaction and providing actionable insights for sellers seeking to improve
product quality. This paper presents a comprehensive benchmarking study comparing a Machine Learning
(ML) approach via the PyCaret AutoML framework against a Deep Learning (DL) approach based on a
Bidirectional Long Short-Term Memory (BiLSTM) architecture with an Attention mechanism for binary
sentiment classification on Indonesian product reviews. The dataset comprises 19,728 samples balanced
equally between positive and negative reviews. For the ML approach, three prominent algorithms were
evaluated via 10-fold stratified cross-validation: Logistic Regression (LR), Support Vector Machine
(SVM) with a linear kernel, and Light Gradient Boosting Machine (LightGBM). Logistic Regression
achieved the best ML performance with an accuracy of 97.26\% and an F1-score of 97.26\%. The BiLSTM
with Attention model, evaluated on 3,946 held-out test samples, achieved an accuracy of 97.24\% and
an F1-score of 97.24\%. These comparative results demonstrate that traditional ML algorithms with
proper preprocessing and feature extraction can compete closely with, and even marginally outperform,
more complex sequential DL architectures on high-dimensional datasets, while simultaneously offering
greater computational efficiency.
\end{abstract}


% ---------------------------------------------------------------
\section{Introduction}
% ---------------------------------------------------------------

The rapid growth of the e-commerce industry in Indonesia has generated an enormous volume of textual
data in the form of product reviews. These reviews contain customer opinions that are highly valuable
for businesses to evaluate their products. However, manually analyzing thousands of reviews is both
inefficient and prone to subjective bias. Consequently, the application of Natural Language Processing
(NLP) techniques for automated sentiment classification has become a crucial solution.

A primary challenge in processing Indonesian product reviews is the prevalent use of colloquial
language, slang (\textit{bahasa gaul}), abbreviations, and irregular sentence structures. Various NLP
studies have compared traditional Machine Learning (ML) and Deep Learning (DL) approaches to address
these challenges. Traditional ML approaches require careful feature engineering, yet the emergence of
AutoML frameworks such as PyCaret~\citep{ali2020pycaret} has substantially streamlined the ML workflow
by automating preprocessing, hyperparameter tuning, and systematic model evaluation at scale.

In contrast, Deep Learning models have demonstrated significant advantages across a range of sequence
processing tasks. Bidirectional Long Short-Term Memory (BiLSTM) networks can learn text representations
from both forward and backward directions to comprehensively retain contextual information. The addition
of an Attention mechanism~\citep{bahdanau2015neural} further enables the model to assign higher weights
to specific tokens that most significantly contribute to the sentiment polarity, regardless of their
position in the sentence.

This study aims to conduct a comprehensive benchmark between leading ML algorithms in a PyCaret AutoML
environment---specifically Logistic Regression, SVM, and LightGBM---against a state-of-the-art DL
architecture (BiLSTM with Attention) implemented using PyTorch. We compare evaluation metrics
including Accuracy, Precision, Recall, and F1-Score to determine which approach is most optimal for
handling a 19,728-sample dataset of Indonesian product reviews. The main contributions of this work
are: (1) a rigorous empirical comparison of ML and DL methods on an Indonesian-language e-commerce
sentiment dataset; (2) a demonstration that classical linear models can remain highly competitive when
paired with appropriate feature engineering; and (3) a fully reproducible experimental pipeline
deployed on Hugging Face Spaces.

% ---------------------------------------------------------------
\section{Related Work}
% ---------------------------------------------------------------

A body of research has examined the effectiveness of Machine Learning and Deep Learning for Indonesian
sentiment analysis. In the traditional ML domain, Support Vector Machines (SVM) and Logistic
Regression have consistently served as strong baselines for text classification due to their robustness
in high-dimensional feature spaces~\citep{riadi2019sentiment,wibowo2020indonesian}. The use of PyCaret
as an AutoML library has gained popularity for its ability to benchmark dozens of models---from LR and
Naive Bayes to SVM, Random Forest, and LightGBM---automatically, thereby minimizing experimental cycle
errors and accelerating prototyping time~\citep{ali2020pycaret,ramadhani2023automl}.

In the Deep Learning domain, RNN-based architectures, particularly LSTM networks introduced by
Hochreiter \& Schmidhuber~\citep{hochreiter1997long}, are widely applied for their capacity to model
long-term dependencies in sequential text. The BiLSTM extension has been shown to further improve
accuracy by observing sequences from two directions~\citep{khasanah2022bilstm}. Additionally, the
integration of the Attention mechanism~\citep{bahdanau2015neural} ensures the model focuses on the
hidden states representing the most sentiment-bearing tokens within a sentence.

Comparative studies between ML and DL for Indonesian sentiment analysis on e-commerce data have shown
mixed results~\citep{purwarianti2019performance,salsabila2022comparison}. LightGBM, introduced by Ke
et al.~\citep{ke2017lightgbm}, has shown competitive performance in tabular and high-dimensional
feature settings, making it a natural candidate for TF-IDF-transformed text features. Our work extends
this line of inquiry by providing a controlled, head-to-head comparison using a modern AutoML pipeline
alongside a custom-built DL model under identical dataset conditions.

% ---------------------------------------------------------------
\section{Dataset}
% ---------------------------------------------------------------

The dataset used in this study originates from customer product reviews collected from Indonesian
marketplace platforms, publicly available on Hugging Face Datasets~\citep{dataset2024}. The total
dataset comprises 19,728 rows with a binary target label (Positive/Negative). The data is split using
a percentage validation technique, yielding 15,782 samples (80\%) for the training set and 3,946
samples (20\%) for the test set. The dataset is near-perfectly balanced, with the test split containing
exactly 1,973 Negative-class reviews and 1,973 Positive-class reviews, ensuring that accuracy metrics
are directly comparable and unbiased by class imbalance.

\subsection{Preprocessing}

The preprocessing pipeline applied to raw text consists of the following sequential steps:

\begin{enumerate}[noitemsep]
  \item \textbf{Case Folding:} All characters are converted to lowercase.
  \item \textbf{Cleansing:} URLs (matching patterns \texttt{http://}, \texttt{https://}, \texttt{www}),
        HTML tags, mentions (\texttt{@user}), hashtags (\texttt{\#topic}), punctuation, and numeric
        characters are removed using regular expressions.
  \item \textbf{Slang Normalization:} Informal Indonesian words are mapped to their standard forms
        using a curated slang dictionary (e.g., \textit{bgt} $\rightarrow$ \textit{banget}). This step
        is critical for handling the high prevalence of colloquial language in user-generated reviews.
  \item \textbf{Tokenization:} Text is split into tokens based on whitespace boundaries. Tokens with
        fewer than 2 characters are subsequently filtered out.
  \item \textbf{Whitespace Normalization:} Redundant whitespace is removed to produce a clean token
        sequence.
\end{enumerate}

\subsection{Feature Extraction for ML}

For the Machine Learning pipeline, cleaned text was transformed into numerical features using TF-IDF
vectorization with the following configuration: \texttt{max\_features=5000},
\texttt{ngram\_range=(1,2)} (unigrams and bigrams), \texttt{min\_df=3}, \texttt{max\_df=0.90}, and
\texttt{sublinear\_tf=True}. This yielded a sparse feature matrix of dimension $N \times 5000$ as
input to the PyCaret classifiers.

% ---------------------------------------------------------------
\section{Methodology}
% ---------------------------------------------------------------

\subsection{Machine Learning Pipeline (PyCaret)}

The PyCaret classification module was used to orchestrate the ML experimental pipeline. The
\texttt{setup()} function was initialized with \texttt{train\_size=0.8} and configured to include
automatic missing-value imputation and MinMax normalization. We focused on benchmarking three
algorithms:

\begin{itemize}[noitemsep]
  \item \textbf{Logistic Regression (LR):} A linear estimator that efficiently maps features to
        discrete class probabilities via a sigmoid function.
  \item \textbf{Support Vector Machine (SVM):} Configured with a linear kernel, trained to maximize
        the separating margin between classes in the high-dimensional TF-IDF feature space.
  \item \textbf{Light Gradient Boosting Machine (LightGBM):} A gradient boosting decision tree
        framework~\citep{ke2017lightgbm} designed for high computational efficiency and low memory
        consumption during training.
\end{itemize}

All models were evaluated using 10-fold Stratified Cross-Validation on the training set, ensuring each
fold maintains the original class distribution.

\subsection{Deep Learning Architecture (BiLSTM + Attention)}

The DL approach was developed entirely within the PyTorch framework, trained from scratch without
pre-trained embeddings. The total parameter count of the model is 1,481,155, comfortably within the
10-million parameter budget. The architecture consists of the following components:

\begin{itemize}[noitemsep]
  \item \textbf{Embedding Layer:} A learnable embedding matrix with a vocabulary size of 13,600,
        mapping tokens to dense vectors of dimension 64.
  \item \textbf{BiLSTM Layer 1:} Input size 64, hidden size 128, bidirectional, yielding an output
        dimension of 256 per timestep.
  \item \textbf{BiLSTM Layer 2:} Input size 256, hidden size 128, bidirectional, maintaining an
        output dimension of 256 per timestep.
  \item \textbf{Attention Layer:} A single linear projection mapping the 256-dimensional feature
        matrix to a scalar attention score per timestep. A softmax operation converts these scores to
        attention weights, and a weighted sum over the BiLSTM output sequence produces a 256-dimensional
        context vector.
  \item \textbf{Classifier Head:} A sequential fully-connected network:
        Linear(256 $\rightarrow$ 64) $\rightarrow$ ReLU $\rightarrow$ Dropout($p$=0.4483)
        $\rightarrow$ Linear(64 $\rightarrow$ 2).
\end{itemize}

% ---------------------------------------------------------------
\section{Experiments}
% ---------------------------------------------------------------

\subsection{Experimental Configuration}

For the ML experiments, the PyCaret \texttt{setup()} automatically handled missing value imputation
and MinMax feature normalization prior to 10-fold cross-validation. For the Deep Learning model,
hyperparameter optimization was conducted over multiple trials; the best configuration (Trial \#8)
yielded: learning rate = 0.0038, weight decay = 0.00107, label smoothing $\approx 7.8 \times 10^{-5}$,
and batch size = 128. The model was trained for up to 10 epochs with early stopping based on validation
loss. The optimal checkpoint was reached at Epoch 5, achieving a best validation loss of 0.0886. All
experiments were executed using Google Colab with GPU acceleration.

\subsection{Evaluation Metrics}

Both ML and DL models are evaluated using Accuracy, Precision, Recall, and F1-Score (macro-averaged).
For the ML models, metrics are averaged across 10 cross-validation folds. For the DL model, metrics
are computed on the held-out test set of 3,946 samples.

% ---------------------------------------------------------------
\section{Results and Discussion}
% ---------------------------------------------------------------

\subsection{ML Results (PyCaret AutoML)}

Table~\ref{tab:ml_results} presents the 10-fold cross-validation results for all three ML models on
the 15,782-sample training set. Logistic Regression emerged as the best-performing model overall.

\begin{table}[htbp]
\centering
\caption{Machine Learning Model Evaluation Results via PyCaret (10-Fold CV, ordered by F1-Score).}
\label{tab:ml_results}
\small
\begin{tabular}{lcccccccc}
\toprule
\textbf{Model} & \textbf{Acc.} & \textbf{AUC} & \textbf{Recall} & \textbf{Prec.} & \textbf{F1}
  & \textbf{Kappa} & \textbf{MCC} & \textbf{Time (s)} \\
\midrule
Logistic Regression  & 0.9726 & 0.9955 & 0.9711 & 0.9741 & 0.9726 & 0.9453 & 0.9453 & 12.130 \\
SVM (Linear Kernel)  & 0.9717 & 0.9946 & 0.9731 & 0.9705 & 0.9718 & 0.9435 & 0.9435 & 6.594  \\
LightGBM             & 0.9643 & 0.9937 & 0.9640 & 0.9645 & 0.9643 & 0.9285 & 0.9286 & 13.408 \\
\bottomrule
\end{tabular}
\end{table}

Logistic Regression achieved an accuracy of 97.26\% and an F1-score of 97.26\%, followed closely by
SVM at 97.17\% and LightGBM at 96.43\%. All three models demonstrated AUC scores above 0.993,
indicating near-perfect discriminative power.

\subsection{DL Results (BiLSTM + Attention)}

The BiLSTM+Attention model was evaluated on the held-out test set of 3,946 samples. The confusion
matrix reveals highly symmetric predictions: True Negatives = 1,918, False Positives = 55, False
Negatives = 54, True Positives = 1,919.

\begin{table}[htbp]
\centering
\caption{Final Evaluation Results of the Deep Learning Model (BiLSTM + Attention) on the Test Set.}
\label{tab:dl_results}
\small
\begin{tabular}{lcccc}
\toprule
\textbf{Class} & \textbf{Precision} & \textbf{Recall} & \textbf{F1-Score} & \textbf{Support} \\
\midrule
Negative (0)        & 0.9726 & 0.9721 & 0.9724 & 1,973 \\
Positive (1)        & 0.9721 & 0.9726 & 0.9724 & 1,973 \\
\midrule
\textbf{Macro Avg}  & \textbf{0.9724} & \textbf{0.9724} & \textbf{0.9724} & \textbf{3,946} \\
\bottomrule
\end{tabular}
\end{table}

\subsection{Benchmark Comparison}

Table~\ref{tab:benchmark} provides the final head-to-head comparison between the best ML model and
the DL architecture.

\begin{table}[htbp]
\centering
\caption{Final Benchmark: ML vs.\ DL Approaches.}
\label{tab:benchmark}
\begin{tabular}{llcc}
\toprule
\textbf{Approach} & \textbf{Best Model} & \textbf{Accuracy} & \textbf{F1-Score} \\
\midrule
Machine Learning (PyCaret)  & Logistic Regression   & 97.26\% & 97.26\% \\
Deep Learning (PyTorch)     & BiLSTM + Attention    & 97.24\% & 97.24\% \\
\bottomrule
\end{tabular}
\end{table}

\begin{figure}[htbp]
  \centering
  \includegraphics[width=0.95\textwidth]{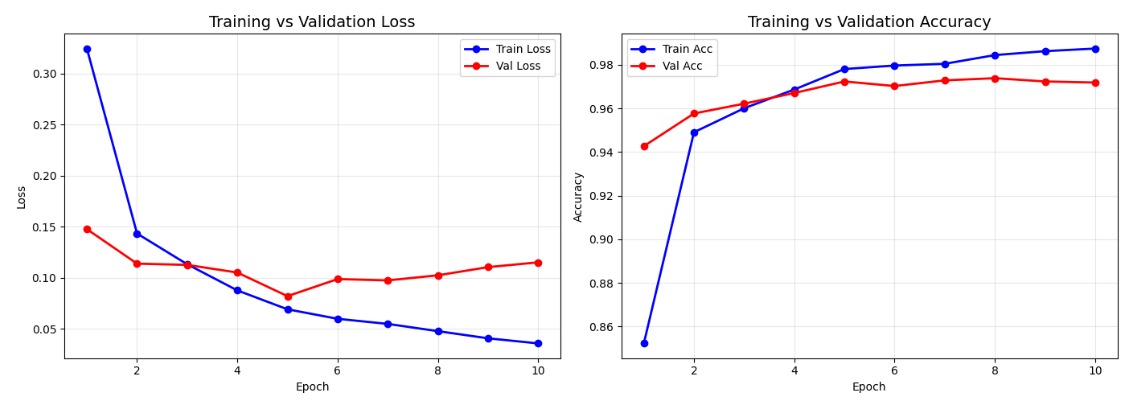}
  \caption{Training and validation loss (left) and accuracy (right) curves for the best BiLSTM+Attention model. The best validation checkpoint is reached at Epoch~5 (Val Loss $\approx 0.082$), after which validation loss begins to increase, indicating the onset of overfitting.}
  \label{fig:training_curves}
\end{figure}

\begin{figure}[htbp]
  \centering
  \includegraphics[width=0.95\textwidth]{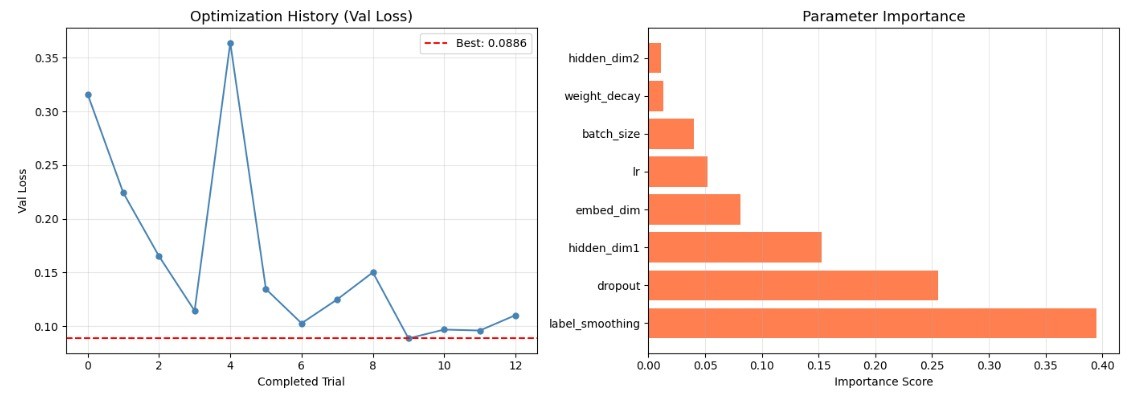}
  \caption{Optuna hyperparameter optimization results. \textit{Left:} Optimization history showing validation loss across 13 trials; the best trial achieves Val Loss = 0.0886 (dashed red line). \textit{Right:} Parameter importance scores, showing that \texttt{label\_smoothing}, \texttt{dropout}, and \texttt{hidden\_dim1} are the most influential hyperparameters.}
  \label{fig:optuna_results}
\end{figure}

\subsection{Discussion}

The benchmark results yield empirically significant and somewhat counterintuitive findings. While it is
commonly assumed that Deep Learning architectures will substantially outperform traditional ML
algorithms on textual data, our experiments on this Indonesian e-commerce review dataset demonstrate
the opposite: the classical linear algorithm Logistic Regression marginally outperforms the
significantly more complex BiLSTM+Attention architecture (97.26\% vs.\ 97.24\% accuracy).

This outcome indicates that the TF-IDF feature representation with 5,000 dimensions produces a feature
space that is already \textit{linearly separable} for this binary classification task. When data is
linearly separable, traditional ML models can achieve near-ceiling performance while offering a
considerable computational advantage---SVM required only approximately 6.5 seconds per fold during
training.

The BiLSTM+Attention model, however, demonstrated notable strengths in classification balance. The
near-identical True Positive and True Negative counts (1,919 vs.\ 1,918) confirm that the Attention
mechanism effectively prevents class bias, enabling robust recognition of both sentiment classes. The
smooth convergence of the training and validation loss curves, with the best checkpoint occurring at
Epoch 5, indicates that the model generalized well without overfitting.

Taken together, these results suggest that for short, high-volume text classification tasks on balanced
datasets, TF-IDF-based ML pipelines via AutoML frameworks like PyCaret provide a highly practical and
competitive alternative to deep sequential models, with the added benefit of interpretability and
deployment simplicity.

% ---------------------------------------------------------------
\section{Conclusion}
% ---------------------------------------------------------------

This study presents a rigorous benchmarking of Machine Learning (Logistic Regression, SVM, LightGBM
via PyCaret AutoML) against Deep Learning (BiLSTM + Attention via PyTorch) for binary sentiment
classification of Indonesian product reviews. The central finding is that the classical Logistic
Regression model (Accuracy: 97.26\%, F1: 97.26\%) marginally outperforms the BiLSTM+Attention
architecture (Accuracy: 97.24\%, F1: 97.24\%), making it the most efficient and effective approach
for this dataset. The ML pipeline via PyCaret is therefore recommended for production inference systems
requiring low computational latency.

Future work may extend this study by evaluating pre-trained Transformer-based models such as IndoBERT
on multi-class sentiment scenarios (positive, neutral, negative), as well as investigating the impact
of more sophisticated Indonesian-specific preprocessing tools such as PySastrawi for stemming.
Additionally, exploring larger and more diverse datasets could reveal conditions under which deep
learning approaches demonstrate their full potential.

% ---------------------------------------------------------------
% References
% ---------------------------------------------------------------
% arXiv note: references.bib is included in the submission.
% arXiv will run BibTeX automatically; a pre-generated .bbl file
% may also be submitted in place of (or alongside) references.bib.
\bibliographystyle{unsrtnat}
\bibliography{references}

\end{document}